\documentclass{nature}

\usepackage{amssymb}
\usepackage{graphicx}
\usepackage{times}
\usepackage{epsfig}
\usepackage{amsmath}
\usepackage{subfigure}
\usepackage{amssymb}
\usepackage{epstopdf}
\graphicspath{ {pic1/} }
\begin{document}

\title{\Large{A LBP Based Correspondence Identification Scheme for Multi-view Sensing Network}}

\author{Raghavendra Kandukuri}

\maketitle
\section{Abstract}
In this paper, we describes a correspondence identification method between two-views of regular RGB camera that can be run in real-time. The basic idea is first applying normalized cross correlation to retrieve a sparse set of matching pairs from image pair. Then loopy belief propagation scheme is applied to the the set of possible candidates to densely identify correspondences from different views. The experiment results demonstrate superb accuracy and precision that outperform the state-of-the-art in the computer vision field. Meanwhile, the implementation is simple enough that can be optimized for real-time performance. We have given the detailed comparison of existing approaches and show that this method can enable various practical applications from 3D reconstruction to image search.

\section{Introduction}
Stereo vision is the most important topic in the computer vision. The key drivers for the most recent sensing system are the efficient 3D geometry estimation that facilitates the visual content creation for many applications from scene completion \cite{app13}.  By and large, stereo calculations can be arranged into two noteworthy classes: neighborhood strategies and worldwide techniques\cite{iref11}. Neighborhood calculations, which depend on connection can have an exceptionally effective execution that is reasonable for ongoing application \cite{iref15}. Worldwide techniques make express smoothness suppositions of the divergence outline limit some cost work. An exemplary class of worldwide techniques is Dynamic Programming \cite{iref1}.  Depth completion can also be achieved by layer depth denoising \cite{cheung13}.

The robotics and computer vision communities have developed many techniques for 3D mapping using range scans [ \cite{iref3}, \cite{iref18}], stereo cameras [\cite{iref23}, \cite{iref15}], monocular cameras \cite{iref5} and even unsorted collection of photos [\cite{iref28}, \cite{iref4}]. Most mapping systems require spatial alignment of consecutive data frames, the detection of loop closure and the globally consistent alignment of all data frames. The objective is not only alignment and registration, but also building 3D models with both shape and appearance information.

\section{Approach Description}
The algorithm can be braked in to two parts: correlation volume computation and hierarchical BP implementation Where D is data term and S is smoothness term \cite{iref1}.

$$E(d)=E_D(d)+E_S(d);$$

\subsection{Correlation volume computation}
In the first figure the pixel X is surrounded by four neighbor's they are Y1,Y2,Y3,Y4 and the smoothness factor between X and all of it neighbor's \cite{iref3}. The next figure shows the updated smoothness factor \cite{iref23}. 
\begin{figure*}[!htb]
\centerline{\epsfig{figure=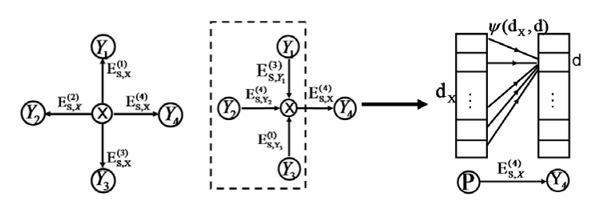, width=11cm}}
\end{figure*}

Step three uses RANSAC to find the best rigid transformation between these feature sets. For sampling the resulting maximum inlier count, the function then effectively computes a more accurate transformation including all inlier points using CATE technique. 

\section{Fast Converging BP}
For standard BP calculations, with a specific end goal to accomplish the best stereo outcomes, an expansive number of cycles are required to ensure the joining \cite{iref28}. Really, there are loads of repetitive calculations required in standard BP. Generally, by just refreshing pixels that have not yet united, quick meeting BP expels those repetitive calculations \cite{iref1}.
In detail, the new smoothness term of one of the pixels in the chart is refreshed by its own particular information term (ED, X), the past smoothness term of its four neighboring pixels (E (i), Old S, Yi), and the single-pass hop cost work  \cite{iref4}.

Figure 2 demonstrates that after a few quantities of cycles, the vast majority of the pixels on the diagram focalize. The quick focalizing BP calculation hence overlooks these pixels, the refreshing plan is just connected to the non-united pixels Figure 2 likewise demonstrates that the running time of the standard BP is direct to the quantity of cycles \cite{iref3} \cite{iref18}. To further enhance accuracy, multiple RGB-D calibration parameters can be applied to step 3 for an optimal estimation \cite{luo14}. 

\begin{figure*}[!htb]
\centerline{\epsfig{figure=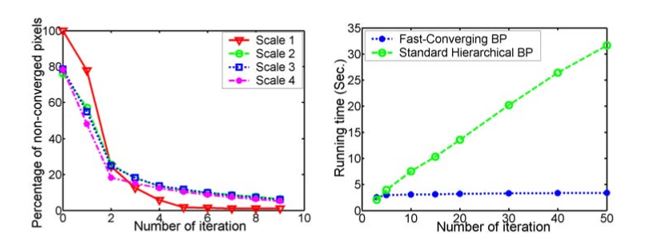, width=11cm}}
\end{figure*}

\section{GPU Implementation}
In our execution, there are four scales in the progressive BP and the principle process for each scale is the same. The refreshing plan for each scale is outlined in Algorithm 1 \cite{iref3}. For each scale, eight surfaces are utilized to store the smoothness term \cite{iref4}.
\begin{figure*}[!htb]
\centerline{\epsfig{figure=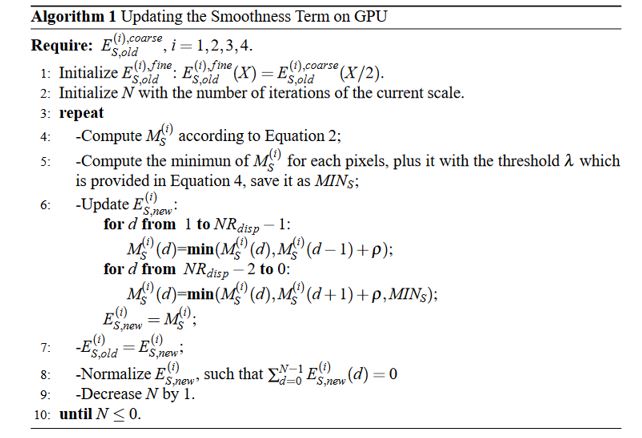, width=11cm}}
\end{figure*}

\section{Experimental Results}
The figures below is from paper \cite{myreference2}. We tried our ongoing BP calculation on a 3 GHz PC running Direct3D 9.0. The GPU is a Geforce 7900 GTX design card with 512M video memory from NVIDIA. All shades are executed utilizing HLSL and arranged utilizing pixel shade 3.0 profile. According to the paper \cite{iref5} \cite{iref28}, the new assessment test information comprises of four stereo matches inside which "Tsukuba" and "Venus" are standard stereo information with inclined surfaces and up to 20 divergence levels, "Teddy" and "Cones" are both recently embraced picture sets with more confounded scene structure and significantly bigger dissimilarity ranges. 
.

Also, as far as exactness, Table 2 demonstrates that our continuous BP beats the various techniques that can accomplish ongoing or close constant execution recorded on the new Middlebury assessment table \cite{tip13}. Since the old assessment table at Middlebury, which contains a few calculations that point continuous, is no longer utilitarian, we gathered the now-impeded mistake rate of the common test information "Tsukuba" and "Venus" and give brings about Table 2 to reference.
\begin{figure*}[!htb]
\centerline{\epsfig{figure=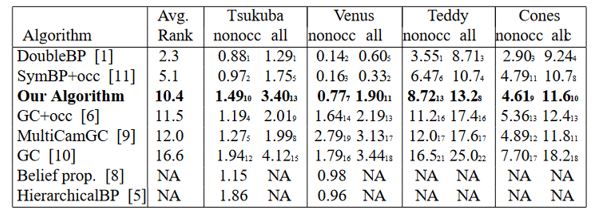, width=11cm}}
\end{figure*}

\subsection{Live System and Efficiency Performance}
\begin{figure*}[!htb]
\centerline{\epsfig{figure=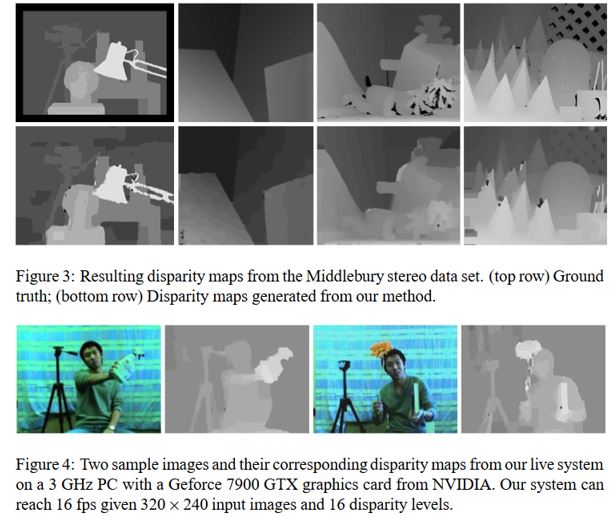, width=11cm}}
\end{figure*}

\section{Conclusion}
In this paper, a continuous stereo model in light of progressive conviction proliferation was proposed, which shows that worldwide advancement based stereo coordinating is workable for Real-time applications. The entire calculation outline in this paper is spotless and brings about top notch stereo coordinating.



\end{document}